\documentclass[a4paper,draftcls,onecolumn,technote,11pt]{IEEEtran}%
\usepackage{amsfonts}
\usepackage{amsmath}
\usepackage{amssymb}
\usepackage{graphicx}%
\providecommand{\U}[1]{\protect \rule{.1in}{.1in}}
\begin{document}

\title{Calibration of Multiple Fish-Eye Cameras Using a Wand}
\author{{\large Qiang Fu$^{1}$, Quan Quan$^{1,2}$, Kai-Yuan Cai$^{1}$}\\1. Department of Automatic Control, Beihang University, Beijing 100191, China\\2. State Key Laboratory of Virtual Reality Technology and Systems, Beihang
University, Beijing 100191, China\\Emails: fq\_buaa@asee.buaa.edu.cn (Qiang Fu); qq\_buaa@buaa.edu.cn (Quan
Quan); kycai@buaa.edu.cn (Kai-Yuan Cai). \thanks{Corresponding author: Qiang
Fu. Email: fq\_buaa@asee.buaa.edu.cn. The camera calibration toolbox is
available at http://quanquan.buaa.edu.cn/.}}
\maketitle

\begin{abstract}
Fish-eye cameras are becoming increasingly popular in computer vision, but
their use for 3D measurement is limited partly due to the lack of an accurate,
efficient and user-friendly calibration procedure. For such a purpose, we
propose a method to calibrate the intrinsic and extrinsic parameters
(including radial distortion parameters) of two/multiple fish-eye cameras
simultaneously by using a wand under general motions. Thanks to the generic
camera model used, the proposed calibration method is also suitable for
two/multiple conventional cameras and mixed cameras (e.g. two conventional
cameras and a fish-eye camera). Simulation and real experiments demonstrate
the effectiveness of the proposed method. Moreover, we develop the camera
calibration toolbox, which is available online.

\end{abstract}

%\author{\large Qiang Fu$^{1}$\\
%\normalsize First School\\
%\normalsize City1\\
%\normalsize Email: first@nodomain.org}


\section{Introduction}

Camera calibration is very important in computer vision, and numerous
researches have been carried out on it. Most of these studies are based on
conventional cameras, which obey the pinhole projection model and provide a
limited overlap region of the field of view (FOV). The overlap region can be
expanded greatly by using fish-eye cameras \cite{Li(2008)}, because fish-eye
cameras can provide images with a very large FOV (about $180^{\circ}$) without
requiring external mirrors or rotating devices \cite{Abraham(2005)}. Fish-eye
cameras have been used in many applications, such as robot navigation
\cite{Shah(1997)}, 3D measurement \cite{Nishimoto(2007)} and city modeling
\cite{Havlena(2008)}. The drawbacks of fish-eye cameras are low resolution and
significant distortion. Their use for 3D measurement is limited partly due to
the lack of an accurate, efficient and user-friendly calibration procedure.

So far, many methods of calibrating conventional cameras \cite{Zhang(2004)}%
,\cite{Heikkila(2000)} have been proposed, but they are inapplicable to
fish-eye camera calibration directly because the pinhole camera model no
longer holds for cameras with a very large FOV. Existing methods of
calibrating fish-eye cameras are roughly classified into three categories: i)
methods based on 3D calibration patterns \cite{Puig(2011)},\cite{Du(2011)},
ii) methods based on 2D calibration patterns \cite{Kannala(2006)}%
,\cite{Mei(2007)},\cite{Feng(2012)}, iii) self-calibration methods
\cite{Micusik(2006)},\cite{Espuny(2011)}. The most widely-used methods are
based on 2D calibration patterns, which are often applicable to a single
camera. In order to calibrate the geometry relation between multiple cameras,
it is required that all cameras observe a sufficient number of points
simultaneously \cite{Zhang(2004)}. It is difficult to achieve by 3D/2D
calibration patterns if two of the cameras face each other. On the other hand,
many wand-based calibration methods \cite{Zhang(2004)},\cite{Franca(2012)}%
,\cite{Pribanic(2009)} were proposed for motion capture systems consisting of
multiple cameras, such as the well-known Vicon system \cite{Vicon}. However,
most of them were dedicated to dealing with conventional cameras. Calibration
methods for fish-eye cameras with a 1D wand have not been discussed in the
literature as far as we know.

For such a purpose, we propose a new method to calibrate the intrinsic and
extrinsic parameters (including radial distortion parameters) of two/multiple
fish-eye cameras simultaneously with a freely-moving wand. Thanks to the
generic camera model used, the proposed calibration method is also suitable
for two/multiple conventional cameras and mixed cameras (e.g. two conventional
cameras and a fish-eye camera). The calibration procedure of two cameras is
summarized as follows. First, the intrinsic and extrinsic parameters are
initialized and optimized by using some prior information such as the real
wand lengths and the nominal focal length provided by the camera manufacturer.
Then, the bundle adjustment \cite{Hartley(2004)} is adopted to refine all
unknowns, which consist of the intrinsic parameters (including radial
distortion parameters), extrinsic parameters and coordinates of 3D points.
With the help of vision graphs in \cite{Kurillo(2008)}, the proposed method is
further extended to the case of multiple cameras, which does not require all
the cameras to have a common FOV. The calibration procedure of multiple
cameras is summarized as follows. First, the intrinsic and extrinsic
parameters of each camera is initialized by involving pairwise calibration
results. Then, the bundle adjustment is used to refine all unknowns, which
consist of the intrinsic and extrinsic parameters (including radial distortion
parameters) of each camera, and coordinates of 3D points.

This paper is organized as follows. Some preliminaries are introduced in
Section II. In Section III, the calibration algorithm for two cameras and
multiple cameras is presented. Then the experimental results are reported in
Section IV, followed by the conclusions in Section V.

\section{Preliminaries}

\subsection{Generic camera model}

The perspective projection is described by the following equation
\cite{Kannala(2006)}:%
\begin{equation}
r_{1}(f,\theta)=f\tan \theta \text{ \  \ (perspective projection)}
\label{perspective}%
\end{equation}
where $\theta$ is the angle between the optical axis and the incoming ray, the
focal length $f$ is fixed for a given camera, and $r_{1}(f,\theta)$ is the
distance between the image point and the principal point. By contrast,
fish-eye lenses are usually designed to obey one of the following projections:%
\begin{align}
r_{2}(f,\theta)  &  =f\theta \text{ \  \  \  \  \  \  \  \  \  \  \  \  \ (equidistance
projection)}\label{equidistance}\\
r_{3}(f,\theta)  &  =f\sin \theta \text{ \  \  \  \  \  \  \  \  \ (orthogonal
projection)}\\
r_{4}(f,\theta)  &  =2f\tan(\theta/2)\text{ \  \  \ (stereographic
projection)}\\
r_{5}(f,\theta)  &  =2f\sin(\theta/2)\text{ \  \  \ (equisolid angle
projection).} \label{equisolid}%
\end{align}
In practice, the real lenses do not satisfy the designed projection model
exactly. A generic camera model for fish-eye lenses is proposed as follows
\cite{Kannala(2006)}%
\begin{equation}
r(\theta)=k_{1}\theta+k_{2}\theta^{3}+k_{3}\theta^{5}+k_{4}\theta^{7}%
+k_{5}\theta^{9}+\cdots. \label{my_model}%
\end{equation}
¡£ \  \ It is found that the first five terms can approximate different
projection curves well. Therefore, in this paper we choose the model that
contains only the five parameters $k_{1},k_{2},k_{3},k_{4},k_{5}.$

As shown in Fig. 1, a 3D point $P$ is imaged at $p$ by a fish-eye camera,
while it would be $p^{\prime}$ by a pinhole camera. Let $O_{c}-X_{c}Y_{c}%
Z_{c}$ denote the camera coordinate system and $o-xy$ the image coordinate
system (unit mm). We can obtain the image coordinates of $p$ in $o-xy$ by%
\begin{equation}
\left(
\begin{array}
[c]{c}%
x\\
y
\end{array}
\right)  =r(\theta)\left(
\begin{array}
[c]{c}%
\cos \varphi \\
\sin \varphi
\end{array}
\right)  \label{my_model2}%
\end{equation}
where $r(\theta)$ is defined in (\ref{my_model}), and $\varphi$ is the angle
between the radial direction and the $x$-axis. Then we can get the pixel
coordinates $\left(  u,v\right)  $ from%
\begin{equation}
\left(
\begin{array}
[c]{c}%
u\\
v
\end{array}
\right)  =\left[
\begin{array}
[c]{cc}%
m_{u} & 0\\
0 & m_{v}%
\end{array}
\right]  \left(
\begin{array}
[c]{c}%
x\\
y
\end{array}
\right)  +\left(
\begin{array}
[c]{c}%
u_{0}\\
v_{0}%
\end{array}
\right)  \label{my_model3}%
\end{equation}
where $\left(  u_{0},v_{0}\right)  $ is the principal point, and $m_{u},m_{v}$
are the number of pixels per unit distance in horizontal and vertical
directions, respectively. Thus, for each fish-eye camera, the intrinsic
parameters are $\left(  k_{1},k_{2},m_{u},m_{v},u_{0},v_{0},k_{3},k_{4}%
,k_{5}\right)  .$

Note that in this paper we do not choose the equivalent sphere model in
\cite{Geyer(2000)}. If this generic model is used, the following calibration
process will not be changed except for some intrinsic parameters. Besides, the
tangential distortion is not considered here for simplicity. As pointed out in
\cite{Kanatani(2013)}, the lens manufacturing technology now is of
sufficiently high levels so that the tangential distortion can be ignored.
Otherwise, the tangential distortion terms need to be taken into account in
(\ref{my_model}). With them, the following calibration process will not be
changed except for some additional unknown parameters.\begin{figure}[h]
\begin{center}
\includegraphics[
scale= 1.0 ]{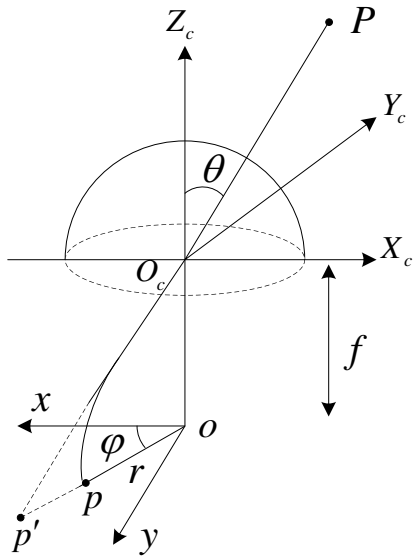} \vspace{-2em}
\end{center}
\caption{Fish-eye camera model [10]. The 3D point $P$ is imaged at $p$ by a
fish-eye camera, while it would be $p^{\prime}$ by a pinhole camera.}%
\end{figure}

\subsection{Essential matrix}

As shown in Fig. 2, the 1D wand has three collinear feature points $A,B,C$
($A_{j},B_{j},C_{j}$ denote their locations for the $j$th image pair), which
satisfy
\[
\left \Vert A-B\right \Vert =L_{1},\left \Vert B-C\right \Vert =L_{2},\left \Vert
A-C\right \Vert =L
\]
where $\left \Vert \mathbf{\cdot}\right \Vert $ denotes the Euclidean vector
norm. Let $O_{0}-X_{0}Y_{0}Z_{0}$ and $O_{1}-X_{1}Y_{1}Z_{1}$ denote the
camera coordinate systems of the left and the right cameras, respectively. The
3D points $A_{j},B_{j},C_{j}$ are projected to $a_{0j},b_{0j},c_{0j}$ on the
unit hemisphere centered at $O_{0}$ and $a_{1j},b_{1j},c_{1j}$ on the unit
hemisphere centered at $O_{1}$. The extrinsic parameters are the rotation
matrix $R\in%
%TCIMACRO{\U{211d} }%
%BeginExpansion
\mathbb{R}
%EndExpansion
^{3\times3}$ and translation vector $T=\left(  t_{x},t_{y},t_{z}\right)
^{T}\in%
%TCIMACRO{\U{211d} }%
%BeginExpansion
\mathbb{R}
%EndExpansion
^{3}$ from the left camera to the right camera. \begin{figure}[h]
\begin{center}
\includegraphics[
scale=1.0]{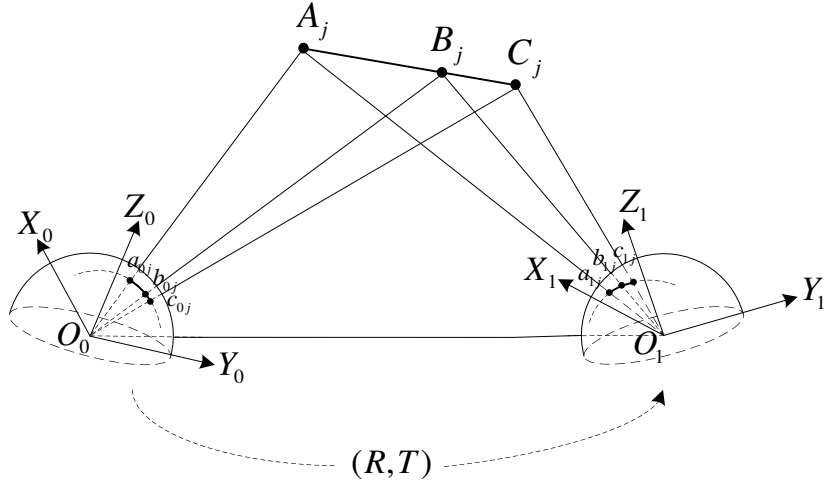}
\end{center}
\par
\vspace{-2em}\caption{Illustration of 1D calibration wand. The 3D points
$A_{j},B_{j},C_{j}$ denote their locations for the $j$th image pair.}%
\end{figure}\

Suppose that a 3D point $M_{j}\in%
%TCIMACRO{\U{211d} }%
%BeginExpansion
\mathbb{R}
%EndExpansion
^{3}$ is projected to
\[
m_{0j}=\left[
\begin{array}
[c]{c}%
\sin \theta_{0j}\cos \varphi_{0j}\\
\sin \theta_{0j}\sin \varphi_{0j}\\
\cos \theta_{0j}%
\end{array}
\right]  ,m_{1j}=\left[
\begin{array}
[c]{c}%
\sin \theta_{1j}\cos \varphi_{1j}\\
\sin \theta_{1j}\sin \varphi_{1j}\\
\cos \theta_{1j}%
\end{array}
\right]
\]
on the unit hemisphere centered at $O_{0}$ and the unit hemisphere centered at
$O_{1},$ respectively. Since $m_{0j},m_{1j},T$ are all coplanar, we have
\cite{Svoboda(1998)}
\begin{equation}
m_{1j}^{T}\left[  T\right]  _{\times}Rm_{0j}=0, \label{rotation3}%
\end{equation}
where
\begin{equation}
\left[  T\right]  _{\times}=\left[
\begin{array}
[c]{ccc}%
0 & -t_{z} & t_{y}\\
t_{z} & 0 & -t_{x}\\
-t_{y} & t_{x} & 0
\end{array}
\right]  .
\end{equation}
Furthermore, (\ref{rotation3}) is rewritten in the form as%
\begin{equation}
m_{1j}^{T}Em_{0j}=0, \label{essential}%
\end{equation}
where $E=\left[  T\right]  _{\times}R$ is known as the \emph{essential matrix}.

\subsection{Reconstruction algorithm}

In this section, a linear reconstruction algorithm for spherical cameras is
proposed, which is the direct analogue of the linear triangulation method for
perspective cameras \cite{Hartley(2004)}. Suppose that the homogeneous
coordinates of a 3D point $M\in%
%TCIMACRO{\U{211d} }%
%BeginExpansion
\mathbb{R}
%EndExpansion
^{3}$ are
\[
M_{0}=\left[
\begin{array}
[c]{c}%
X_{0}\\
Y_{0}\\
Z_{0}\\
1
\end{array}
\right]  ,M_{1}=\left[
\begin{array}
[c]{c}%
X_{1}\\
Y_{1}\\
Z_{1}\\
1
\end{array}
\right]  =[R,T]M_{0}%
\]
in $O_{0}-X_{0}Y_{0}Z_{0}$ and $O_{1}-X_{1}Y_{1}Z_{1}$, respectively. The 3D
point $M$ is projected to%
\[
m_{0}=\left[
\begin{array}
[c]{c}%
\sin \theta_{0}\cos \varphi_{0}\\
\sin \theta_{0}\sin \varphi_{0}\\
\cos \theta_{0}%
\end{array}
\right]  ,m_{1}=\left[
\begin{array}
[c]{c}%
\sin \theta_{1}\cos \varphi_{1}\\
\sin \theta_{1}\sin \varphi_{1}\\
\cos \theta_{1}%
\end{array}
\right]
\]
on the unit hemisphere centered at $O_{0}$ and the unit hemisphere centered at
$O_{1}$, respectively. Then we have%
\begin{equation}
\left \{
\begin{array}
[c]{c}%
s_{0}m_{0}=Q_{0}M_{0}\\
s_{1}m_{1}=Q_{1}M_{1}%
\end{array}
\right.
\end{equation}
where $s_{0},s_{1}$ are scale factors and $Q_{0}=\left[  I_{3},0_{3\times
1}\right]  \in%
%TCIMACRO{\U{211d} }%
%BeginExpansion
\mathbb{R}
%EndExpansion
^{3\times4},$ $Q_{1}=\left[  R,T\right]  \in%
%TCIMACRO{\U{211d} }%
%BeginExpansion
\mathbb{R}
%EndExpansion
^{3\times4}.$ For each image point on the unit hemisphere, the scale factor
can be eliminated by a cross product to give three equations, two of which are
linearly independent. So the four independent equations are written in the
form as follows
\begin{equation}
AM_{0}=0, \label{rebuild6}%
\end{equation}
with%
\begin{equation}
A=\left[
\begin{array}
[c]{c}%
\sin \theta_{0}\cos \varphi_{0}Q_{0,3}-\cos \theta_{0}Q_{0,1}\\
\sin \theta_{0}\sin \varphi_{0}Q_{0,3}-\cos \theta_{0}Q_{0,2}\\
\sin \theta_{1}\cos \varphi_{1}Q_{1,3}-\cos \theta_{1}Q_{1,1}\\
\sin \theta_{1}\sin \varphi_{1}Q_{1,3}-\cos \theta_{1}Q_{1,2}%
\end{array}
\right]  ,
\end{equation}
where $Q_{0,i}$ and $Q_{1,i}$ are the $i$th row of $Q_{0}$ and $Q_{1}$,
respectively. Based on (\ref{rebuild6}), $M_{0}$ is the singular vector
corresponding to the smallest singular value of $A$. So far, given
$m_{0},m_{1},R,T,$ the homogeneous coordinates of $M\in%
%TCIMACRO{\U{211d} }%
%BeginExpansion
\mathbb{R}
%EndExpansion
^{3}$ in $O_{0}-X_{0}Y_{0}Z_{0}$, namely $M_{0},$ is reconstructed. This is
called \emph{the linear reconstruction algorithm}.

Note that equation (\ref{rebuild6}) provides only a linear solution, which is
not very accurate in presence of noises. It could be refined by minimizing
reprojection errors or Sampson errors \cite{Hartley(2004)}. However, since the
reconstruction algorithm is carried out at each optimization iteration, it is
more efficient to choose the linear reconstruction algorithm mentioned above.
Furthermore, the linear reconstruction algorithm can be extended easily to the
case of $n$-view ($n>2$) triangulation for calibration of multiple cameras
(section III-B) \cite{Hartley(2004)}.

\section{Calibration algorithm}

\subsection{Calibration of two cameras}

Based on the preliminaries mentioned in section II, we next present a generic
method to simultaneously calibrate the intrinsic and extrinsic parameters
(including radial distortion parameters) of two cameras with a freely-moving
1D wand, which contains three points in known positions, as shown in Fig. 7
(a). This method is simple, user-friendly and can be used to calibrate two
fish-eye cameras. Let the intrinsic parameters of the $i$th camera be
$(k_{1}^{i},k_{2}^{i},m_{u}^{i},m_{v}^{i},u_{0}^{i},v_{0}^{i},k_{3}^{i}%
,k_{4}^{i},k_{5}^{i})$. Without loss of generality, we take the $0$th camera
and $1$th camera as an example in this subsection. The first three steps of
the calibration procedure involve only twelve intrinsic parameters $(k_{1}%
^{0},k_{2}^{0},m_{u}^{0},m_{v}^{0},u_{0}^{0},v_{0}^{0},k_{1}^{1},k_{2}%
^{1},m_{u}^{1},m_{v}^{1},u_{0}^{1},v_{0}^{1})$, leaving the other parameters
dealt with only in the final step.

\textbf{Step 1: Initialization of intrinsic parameters}. For the $i$th camera,
the principal point $\left(  u_{0}^{i},v_{0}^{i}\right)  $ is initialized by
the coordinates of the image center, and the pixel sizes $m_{u}^{i}$ and
$m_{v}^{i}$ are given by the camera manufacturer. If the $i$th camera is a
conventional or fish-eye camera, then the initial values of $k^{i}=\left(
k_{1}^{i},k_{2}^{i}\right)  ^{T}$ are obtained by fitting the model
(\ref{my_model}) to the projections (\ref{perspective})-(\ref{equisolid}).
Concretely, let the interval $\left[  0,\theta_{\max}^{i}\right]  $ be equally
divided into many pieces $\left[  \theta_{1}^{i},\theta_{2}^{i},\cdots
,\theta_{p}^{i}\right]  \in%
%TCIMACRO{\U{211d} }%
%BeginExpansion
\mathbb{R}
%EndExpansion
^{p}.$ Then we have%
\begin{equation}
\left[
\begin{array}
[c]{cc}%
\theta_{1}^{i} & \theta_{1}^{i3}\\
\theta_{2}^{i} & \theta_{2}^{i3}\\
\vdots & \vdots \\
\theta_{p}^{i} & \theta_{p}^{i3}%
\end{array}
\right]  \left[
\begin{array}
[c]{c}%
k_{1,s}^{i}\\
k_{2,s}^{i}%
\end{array}
\right]  =\left[
\begin{array}
[c]{c}%
r_{s}(f^{i},\theta_{1}^{i})\\
r_{s}(f^{i},\theta_{2}^{i})\\
\vdots \\
r_{s}(f^{i},\theta_{p}^{i})
\end{array}
\right]  ,\text{ }s=1,2,\cdots,5.\label{fitmodel}%
\end{equation}
where the nominal focal length of the $i$th camera is $f^{i}$ and the maximum
viewing angle is $\theta_{\max}^{i}$ provided by the camera manufacturer.
Based on (\ref{fitmodel}), for the $i$th camera, $k^{i}$ is determined by
\begin{equation}
\left(  k^{i},s^{\ast}\right)  =\arg \underset{k_{1,s}^{i},k_{2,s}^{i}%
,s\in \left \{  1,2,\cdots,5\right \}  }{\min}\sum_{j=1}^{p}\left(  r_{s}%
(f^{i},\theta_{j}^{i})-k_{1,s}^{i}\theta_{j}^{i}-k_{2,s}^{i}\theta_{j}%
^{i3}\right)  ^{2}.\label{fitobject}%
\end{equation}
So far, we get the initialization of intrinsic parameters $(k_{1}^{i}%
,k_{2}^{i},m_{u}^{i},m_{v}^{i},u_{0}^{i},v_{0}^{i}),i=0,1$. Note that it is
required to specify the projection type of cameras in advance in
\cite{Kannala(2006)}. Otherwise, it is possible to get inaccurate calibration
results. However, this is not a problem in this paper because we obtain the
best initialization of $k^{i}$ automatically. Besides this, the initialization
of the principle point is reasonable, because the principal point of modern
digital cameras lies close to the center of the image \cite{Hartley(2004)}.

\textbf{Step 2: Initialization of extrinsic parameters. }With the intrinsic
parameters $(k_{1}^{i},k_{2}^{i},m_{u}^{i},m_{v}^{i},u_{0}^{i},v_{0}%
^{i}),$\\$i=0,1$ and the pixel coordinates of image points for the
$j$th image pair, we can compute
$\theta_{0j},\varphi_{0j},\theta_{1j}$ and $\varphi_{1j}$ by
(\ref{my_model})-(\ref{my_model3}). Therefore, according to
(\ref{essential}), the essential matrix $E_{01}$ is obtained by
using the 5-point random sample consensus (RANSAC) algorithm
\cite{Nister(2004)} if five or more corresponding points are given.

If the essential matrix $E_{01}$ is known, then the initial values for the
extrinsic parameters $R_{01}$ and $\bar{T}_{01}$ are obtained by the singular
value decomposition of $E_{01}$ \cite{Hartley(2004)}. Note that $\left \Vert
\bar{T}_{01}\right \Vert =1$, so the obtained translation vector $\bar{T}_{01}$
differs from the real translation vector $T_{01}$ by a scale factor. Let
$A_{j}^{r},C_{j}^{r}$ denote the reconstructed points of $A,C$ for the $j$th
image pair, which are given by the linear reconstruction algorithm based on
(\ref{rebuild6}) with the intrinsic and extrinsic parameters obtained above.
In order to minimize errors, the scale factor $\lambda$ is%
\begin{equation}
\lambda=\frac{1}{N}\sum_{j=1}^{N}\frac{L}{\left \Vert A_{j}^{r}-C_{j}%
^{r}\right \Vert },
\end{equation}
where $N$ is the number of image pairs. Finally, the initial value for the
translation vector is%
\begin{equation}
T_{01}=\left(  t_{x},t_{y},t_{z}\right)  ^{T}=\lambda \bar{T}_{01}\in%
%TCIMACRO{\U{211d} }%
%BeginExpansion
\mathbb{R}
%EndExpansion
^{3}.
\end{equation}
Thus, we obtain the initialization of extrinsic parameters $R_{01}$ and
$T_{01}$.

\textbf{Step 3: Nonlinear optimization of intrinsic and extrinsic parameters.
}Denote\textbf{ }the reconstructed points of $A,B,C$ for the $j$th image pair
by $A_{j}^{r},B_{j}^{r},C_{j}^{r}$ respectively, which are given by the linear
reconstruction algorithm based on (\ref{rebuild6}) with the intrinsic and
extrinsic parameters obtained above. Because of noises, there exist distance
errors as follows%
\begin{align}
g_{1,j}\left(  x\right)   &  =L_{1}-\left \Vert A_{j}^{r}-B_{j}^{r}\right \Vert
\label{object1}\\
g_{2,j}\left(  x\right)   &  =L_{2}-\left \Vert B_{j}^{r}-C_{j}^{r}\right \Vert
\label{object2}\\
g_{3,j}\left(  x\right)   &  =L-\left \Vert A_{j}^{r}-C_{j}^{r}\right \Vert
\label{object3}%
\end{align}
where $x=\left(  k_{1}^{0},k_{2}^{0},m_{u}^{0},m_{v}^{0},u_{0}^{0},v_{0}%
^{0},k_{1}^{1},k_{2}^{1},m_{u}^{1},m_{v}^{1},u_{0}^{1},v_{0}^{1},r_{1}%
,r_{2},r_{3},t_{x},t_{y},t_{z}\right)  \in%
%TCIMACRO{\U{211d} }%
%BeginExpansion
\mathbb{R}
%EndExpansion
^{18}.$ In particular, $r_{01}=(r_{1},r_{2},r_{3})^{T}$ $\in%
%TCIMACRO{\U{211d} }%
%BeginExpansion
\mathbb{R}
%EndExpansion
^{3}$ and the rotation matrix $R_{01}$ are related by the Rodrigues formula,
namely $R_{01}=e^{\left[  r_{01}\right]  _{\times}}$ \cite[p. 585]%
{Hartley(2004)}. Therefore, according to equations (\ref{object1}%
)-(\ref{object3}), the objective function for optimization is%
\begin{equation}
x^{\ast}=\arg \underset{x}{\min}\sum_{j=1}^{N}\left(  g_{1,j}^{2}\left(
x\right)  +g_{2,j}^{2}\left(  x\right)  +g_{3,j}^{2}\left(  x\right)  \right)
,
\end{equation}
which is solved by using the Levenberg-Marquardt method \cite{Hartley(2004)}.

\textbf{Step 4: Bundle adjustment. }The solution above can be refined through
the bundle adjustment \cite{Hartley(2004)}, which involves both the camera
parameters and 3D space points. For the $j$th image pair, we can compute
$A_{j}^{r},B_{j}^{r},C_{j}^{r}$ by the linear reconstruction algorithm based
on equation (\ref{rebuild6}) with the camera parameters $x^{\ast}$ obtained in
\textit{Step 3}\textbf{. }If\textbf{ }%
\[
\left \vert \frac{L-\left \Vert A_{j}^{r}-C_{j}^{r}\right \Vert }{L}\right \vert
>1\%
\]
then the $j$th image pair is removed from the observations. After this, the
number of image pairs reduces from $N$ to $N_{1}$. Without loss of generality,
the image pairs from $\left(  N_{1}+1\right)  $th to $N$th are removed. Since
the 3D space points $A_{j},B_{j}$ and $C_{j}$ are collinear, they have the
relation as follows%
\begin{equation}
\left \{
\begin{array}
[c]{c}%
B_{j}=f_{B}(A_{j},\phi_{j},\theta_{j})=A_{j}+L_{1}\cdot n_{j}\\
C_{j}=f_{C}(A_{j},\phi_{j},\theta_{j})=A_{j}+L\cdot n_{j}%
\end{array}
\right.  , \label{orientation}%
\end{equation}
where $\phi_{j},\theta_{j}$ are spherical coordinates centered at $A_{j}$ and
$n_{j}=\left[  \sin \phi_{j}\cos \theta_{j},\sin \phi_{j}\sin \theta_{j},\cos
\phi_{j}\right]  ^{T}$ denotes the orientation of the 1D wand.

The six additional camera parameters $\left(  k_{3}^{0},k_{4}^{0},k_{5}%
^{0},k_{3}^{1},k_{4}^{1},k_{5}^{1}\right)  $ for the two cameras are
initialized to zero first, which together with $x^{\ast}$ constitute%
\[
y=\left(  k_{1}^{0},k_{2}^{0},m_{u}^{0},m_{v}^{0},u_{0}^{0},v_{0}^{0}%
,k_{3}^{0},k_{4}^{0},k_{5}^{0},k_{1}^{1},k_{2}^{1},m_{u}^{1},m_{v}^{1}%
,u_{0}^{1},v_{0}^{1},k_{3}^{1},k_{4}^{1},k_{5}^{1},r_{1},r_{2},r_{3}%
,t_{x},t_{y},t_{z}\right)  \in%
%TCIMACRO{\U{211d} }%
%BeginExpansion
\mathbb{R}
%EndExpansion
^{24}.
\]
Let functions $P_{i}(y,M)$ denote the projection of a 3D point $M$ onto the
$i$th camera image plane under the parameter $y,$ $i=0,1$. Bundle adjustment
minimizes the following reprojection error%
\begin{equation}
\underset{y,A_{j},\phi_{j},\theta_{j}}{\min}\text{ }\sum_{i=0}^{1}\sum
_{j=1}^{N_{1}}\left(  \left \Vert a_{ij}-P_{i}\left(  y,A_{j}\right)
\right \Vert ^{2}+\left \Vert b_{ij}-P_{i}\left(  y,f_{B}(A_{j},\phi_{j}%
,\theta_{j})\right)  \right \Vert ^{2}+\left \Vert c_{ij}-P_{i}\left(
y,f_{C}(A_{j},\phi_{j},\theta_{j})\right)  \right \Vert ^{2}\right)
\end{equation}
where $a_{ij},b_{ij},c_{ij}$ are the image points of 3D points $A_{j}%
,B_{j},C_{j}$ in the $i$th camera respectively. Since $A_{j}^{r},B_{j}%
^{r},C_{j}^{r}$ are known, we could obtain $\phi_{j}^{r},\theta_{j}^{r}$ from
(\ref{orientation}). Then, $A_{j},\phi_{j},\theta_{j}$ are initialized by
$A_{j}^{r},\phi_{j}^{r},\theta_{j}^{r}$ respectively. After all the
optimization variables are initialized, the nonlinear minimization is done
using the Sparse Levenberg-Marquardt algorithm \cite{Lourakis(2010)}.

Note that the main difference here from existing work is to take the extra
parameters of the radial distortion in the set of unknowns into bundle adjustment.

\subsection{Calibration of multiple cameras}

\textbf{Step 1: Initialization of intrinsic and extrinsic parameters}. The
multiple camera system could be represented by a weighted undirected graph as
in \cite{Kurillo(2008)}. For example, the vision graph of a system consisting
of five cameras is shown in Fig. 3. Each vertex represents an individual
camera and the weights $w_{ij}$ are given as $\frac{1}{M_{ij}}$ where $M_{ij}$
is the number of points in the common field of view of the two cameras. If
$M_{ij}=0$, then the vertices corresponding to the two cameras are not
connected. Next, we use the Dijkstra's shortest path algorithm
\cite{Chen(2003)} to find the optimal path from a reference camera to other
cameras. With the shortest paths from the reference camera to other cameras
and corresponding pairwise calibration results, we could get the rotation
matrices and translation vectors that represent the transformation from the
reference camera to other cameras. For example, if the transformations from
the $i$th camera to $j$th camera and from the $j$th camera to $k$th camera are
$(R_{ij},T_{ij})$ and $(R_{jk},T_{jk})$ respectively, then the transformation
from the $i$th camera to $k$th camera is obtained as follows:%
\begin{equation}
\left \{
\begin{array}
[c]{l}%
R_{ik}=R_{jk}R_{ij}\\
T_{ik}=R_{jk}T_{ij}+T_{jk}%
\end{array}
\right.  . \label{transfer}%
\end{equation}
\begin{figure}[h]
\begin{center}
\includegraphics[
scale=0.85]{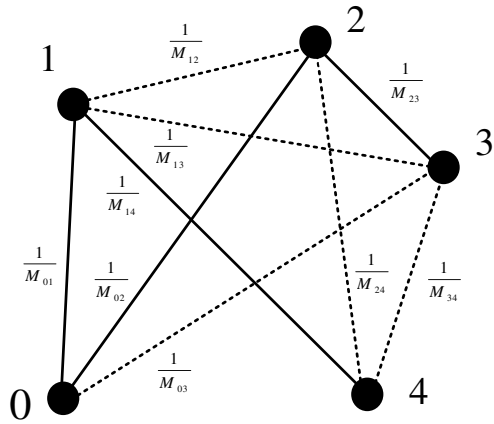}
\end{center}
\par
\vspace{-2em}\caption{Vision graph and the optimal path from reference camera
0 to the other four cameras in solid lines. $M_{ij}$ is the number of common
points between cameras and $\frac{1}{M_{ij}}$ is the corresponding weight.
Vertices 0 and 4 are not connected because $M_{04}=0$.}%
\end{figure}If the length of a path from the reference camera is longer than
two, we could apply the equation (\ref{transfer}) sequentially to cover the
entire path. Besides, the initial value of each camera's intrinsic parameters
is determined from the corresponding pairwise calibration results when the
most points exist in the common field of view of two cameras.

Note that only the pairwise calibration involved in the optimal path is
performed by using the calibration algorithm of two cameras mentioned before.
However, if all the camera pairs are calibrated as in \cite{Kurillo(2008)},
then it will be very time-consuming especially when the number of cameras is large.

\textbf{Step 2: Bundle adjustment.} As in the calibration algorithm of two
cameras, $A_{j}^{r},B_{j}^{r},C_{j}^{r}$ are computed by $n$-view ($n\geq2$)
triangulation method in section II-C and a distance error threshold can be set
to remove outliers. The intrinsic and extrinsic parameters of $m+1$ cameras
(except the extrinsic parameter of the reference camera---the $0$th camera, as
it is constantly $I_{3\times3}$ and $0_{3\times1}$) constitute $y\in%
%TCIMACRO{\U{211d} }%
%BeginExpansion
\mathbb{R}
%EndExpansion
^{15m+9}$. Let functions $P_{i}(y,M)$ $(i=0,1,\cdots,m)$ define projection of
a 3D point $M$ onto the $i$th camera image plane, then bundle adjustment
minimizes the following reprojection error%
\begin{equation}
\underset{y,A_{j},\phi_{j},\theta_{j}}{\min}\text{ }\sum_{i=0}^{m}\sum
_{j=1}^{N_{i}}\left(  \left \Vert a_{ij}-P_{i}\left(  y,A_{j}\right)
\right \Vert ^{2}+\left \Vert b_{ij}-P_{i}\left(  y,f_{B}(A_{j},\phi_{j}%
,\theta_{j})\right)  \right \Vert ^{2}+\left \Vert c_{ij}-P_{i}\left(
y,f_{C}(A_{j},\phi_{j},\theta_{j})\right)  \right \Vert ^{2}\right)
\end{equation}
where $a_{ij},b_{ij},c_{ij}$ are the image points of 3D points $A_{j}%
,B_{j},C_{j}$ in the $i$th camera, and $N_{i}$ is the number of times
$A_{j},B_{j},C_{j}$ are viewed in the $i$th camera. After all the optimization
variables are initialized, the nonlinear minimization is done by using the
Sparse Levenberg-Marquardt algorithm \cite{Lourakis(2010)}.

\section{Experimental results}

\subsection{Simulation experiments}

\subsubsection{Simulation setting}

In the simulation experiments, the $0$th$,1$th$,2$th fish-eye cameras all have
image resolutions of 640 pixels $\times$ 480 pixels with pixel sizes of
$5.6\mu m\times5.6\mu m$ and FOVs of $185^{\circ}$. As for the 1D calibration
wand, the feature points $A$ and $B,C$ satisfy%
\begin{align*}
L_{1}  &  =\left \Vert A-B\right \Vert =400mm\\
L_{2}  &  =\left \Vert B-C\right \Vert =200mm\\
L  &  =\left \Vert A-C\right \Vert =600mm.
\end{align*}
Suppose that the 1D calibration wand undertake 300 times with
general motions inside the volume of
$\left[-0.35,0.35\right]$m$\times \left[-0.35,0.35\right]$m$\times
\left[0.7,1\right]$m. The rotation matrices from the $0$th to the
$1$th$,2$th cameras are $\left[ 28.65,28.65,28.65\right]  ^{T}$,
$\left[ 57.3,57.3,57.3\right] ^{T}$ (in the form of Euler angles,
unit: degree), respectively. The translation vectors from the $0$th
to the $1$th$,2$th cameras are $\left[  -700,100,200\right]
^{T},\left[  -1200,-200,700\right] ^{T}$, respectively. The
calibration error of rotation is measured by the absolute error in
degrees between the true rotation matrix $R_{true}$ and the
estimated rotation matrix $R$ defined as \cite{Zheng(2013)}
\begin{equation}
E_{r}=\text{max}_{k=1}^{3}\left \Vert \text{acos}\left \langle r_{true}%
^{k},r^{k}\right \rangle \right \Vert \times180/\pi,
\end{equation}
where $r_{true}^{k}$ and $r^{k}$ are the $k$th column of $R_{true}$ and $R$,
respectively. The calibration error of translation is measured by%
\begin{equation}
E_{t}=\frac{\left \Vert T_{true}-T\right \Vert }{\left \Vert T_{true}\right \Vert
}.
\end{equation}
where the true translation vector is $T_{true}$ and the estimated translation
vector is $T$. If there are $n$ 3D points viewed by a camera, the global
calibration accuracy of this camera is evaluated by the root-mean-squared
(RMS) reprojection error%
\begin{equation}
E_{RMS}=\sqrt{\frac{1}{n}\sum_{j=1}^{n}\left(  u_{j}-\hat{u}_{j}\right)  ^{2}%
},
\end{equation}
where $u_{j}$ denotes the image point of the $j$th 3D point and $\hat{u}_{j}$
is the corresponding reprojection point obtained by using calibration results.
Next, we perform simulation for both two cameras (the $0$th$,1$th fish-eye
cameras) and multiple cameras (the $0$th$,1$th$,2$th fish-eye cameras).

\subsubsection{Noise simulations}

The truth values of the three cameras' focal lengths and principal points are
2 mm and $\left(  310,250\right)  $, while initial values are 1.8 mm and
$\left(  320,240\right)  $, respectively. Gaussian noises with the mean value
$\mu=0$ and the standard deviation $\sigma$ varying from 0 to 2 pixels are
added to the image points. Simulations are performed 10 times for each noise
level and the average of estimated parameters is taken as the result. Fig. 4
(a)-(e) show the calibration errors of the intrinsic and extrinsic parameters,
while Fig. 4 (f) gives the RMS reprojection errors of the cameras. In Fig. 4,
`2cams' means calibration of the $0$th$,1$th fish-eye cameras (two cameras),
and `3cams' means calibration of the $0$th$,1$th$,2$th fish-eye cameras
(multiple cameras).\begin{figure}[h]
\begin{center}
\includegraphics[
scale=0.7]{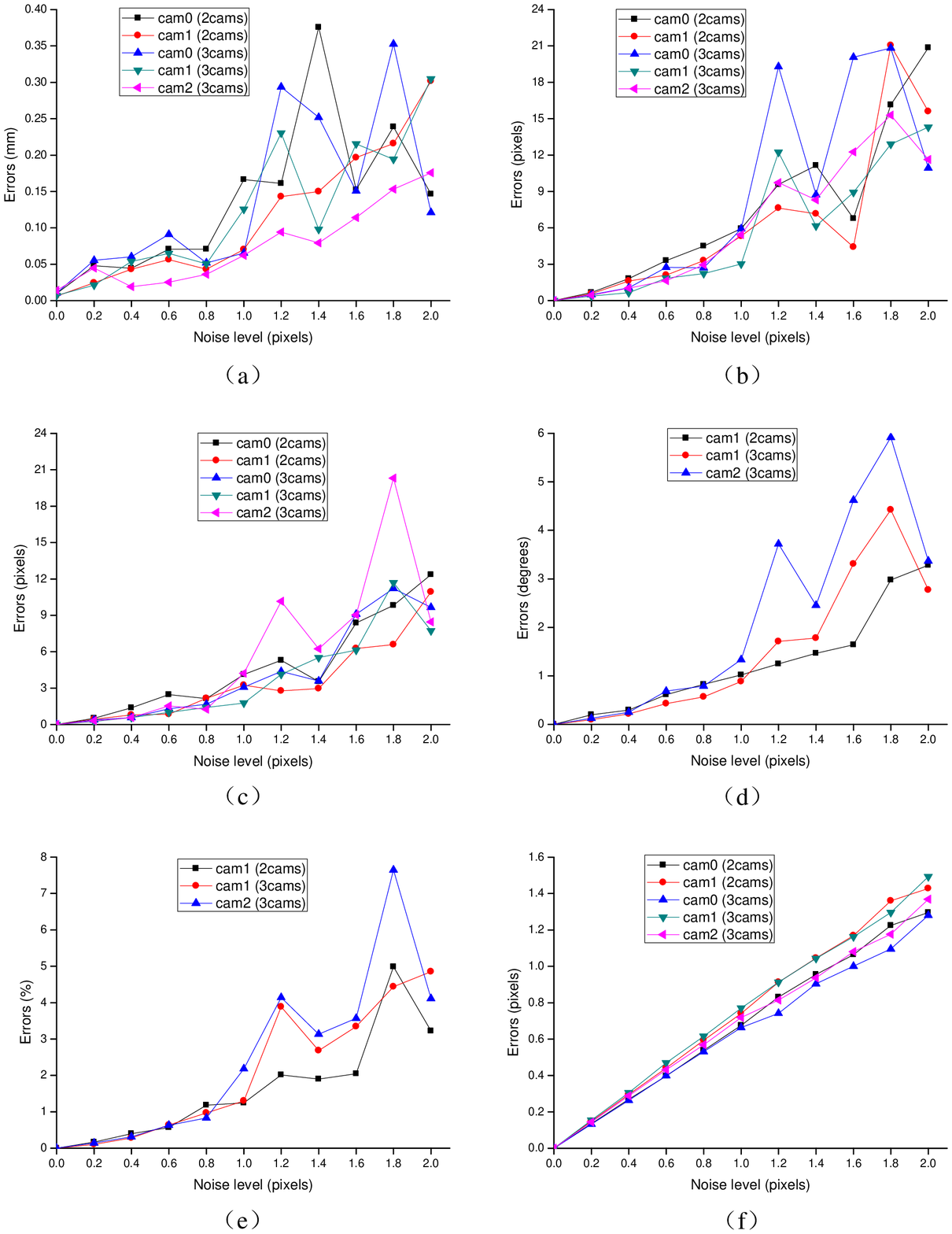}
\end{center}
\par
\vspace{-2em}\caption{Calibration errors of the intrinsic and extrinsic
parameters and reprojection errors for different noise levels. `2cams' means
calibration of the $0th,1th$ fish-eye cameras and `3cams' means calibration of
the $0th,1th,2th$ fish-eye cameras: (a) focal lengths of two and multiple
cameras; (b) principle points ($u_{0}$) of two and multiple cameras; (c)
principle points ($v_{0}$) of two and multiple cameras; (d) calibration error
of rotation $E_{r}$ of two and multiple cameras; (e) calibration error of
translation $E_{t}$ of two and multiple cameras; (f) RMS reprojection error
$E_{RMS}$ of two and multiple cameras.}%
\end{figure}

As shown in Fig. 4, the calibration errors of the intrinsic and extrinsic
parameters do not change drastically with the noise level. Moreover, the RMS
reprojection errors of the cameras increase almost linearly with the noise
level. All these errors are small even when $\sigma=2$ pixels. This shows that
the calibration algorithm in this paper performs well and achieves high
stability for the cases of both two cameras and multiple cameras.

\subsubsection{Initial value simulations}

The truth values of the three cameras' principal points are $\left(
310,250\right)  $, while initial values are $\left(  320,240\right)  $.
Gaussian noise with the mean value $\mu=0$ and the standard deviation
$\sigma=1$ pixel are added to the image points. The truth values of the three
cameras' focal lengths vary from 1.5 mm to 2.5 mm, while the initial values
are fixed to 2 mm. Simulations are performed 10 times for each focal length
and the average of involving parameters is taken as the result. Fig. 5 (a)-(e)
show the calibration errors of the intrinsic, while Fig. 5 (f) gives the
extrinsic parameters and the RMS reprojection errors of the
cameras.\begin{figure}[h]
\begin{center}
\includegraphics[
scale=0.7]{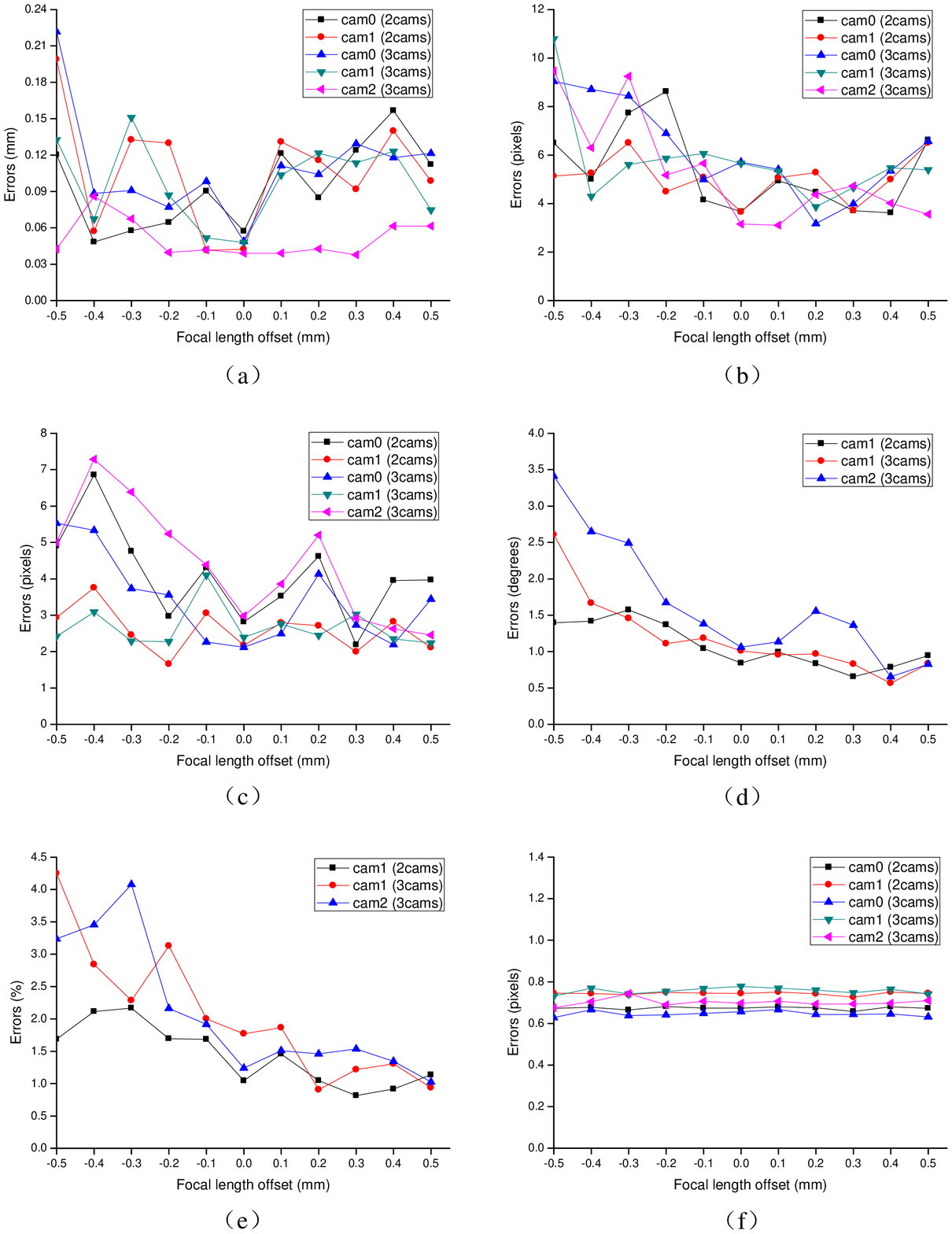}
\end{center}
\par
\vspace{-2em}\caption{Calibration errors of the intrinsic and extrinsic
parameters and reprojection errors for different focal length offsets. `2cams'
means calibration of the $0th,1th$ fish-eye cameras and `3cams' means
calibration of the $0th,1th,2th$ fish-eye cameras: (a) focal lengths of two
and multiple cameras; (b) principle points ($u_{0}$) of two and multiple
cameras; (c) principle points ($v_{0}$) of two and multiple cameras; (d)
calibration error of rotation $E_{r}$ of two and multiple cameras; (e)
calibration error of translation $E_{t}$ of two and multiple cameras; (f) RMS
reprojection error $E_{RMS}$ of two and multiple cameras.}%
\end{figure}

Next, the true values of the three cameras' focal lengths are 2 mm, while the
initial values are 1.8 mm. Gaussian noise with the mean value $\mu=0$ and the
standard deviation $\sigma=1$ pixel is added to the image points. The truth
values of the three cameras' principle points vary from $\left(
270,190\right)  $ to $\left(  370,290\right)  $ along the diagonal line $u=v$,
while the initial values are fixed to $\left(  320,240\right)  $. Experiments
are performed 10 times for each principle point and the average of estimated
parameters is taken as the result. Fig. 6 (a)-(e) show the calibration errors
of the intrinsic and extrinsic parameters, while Fig. 6 (f) gives the RMS
reprojection errors of the cameras.\begin{figure}[h]
\begin{center}
\includegraphics[
scale=0.7]{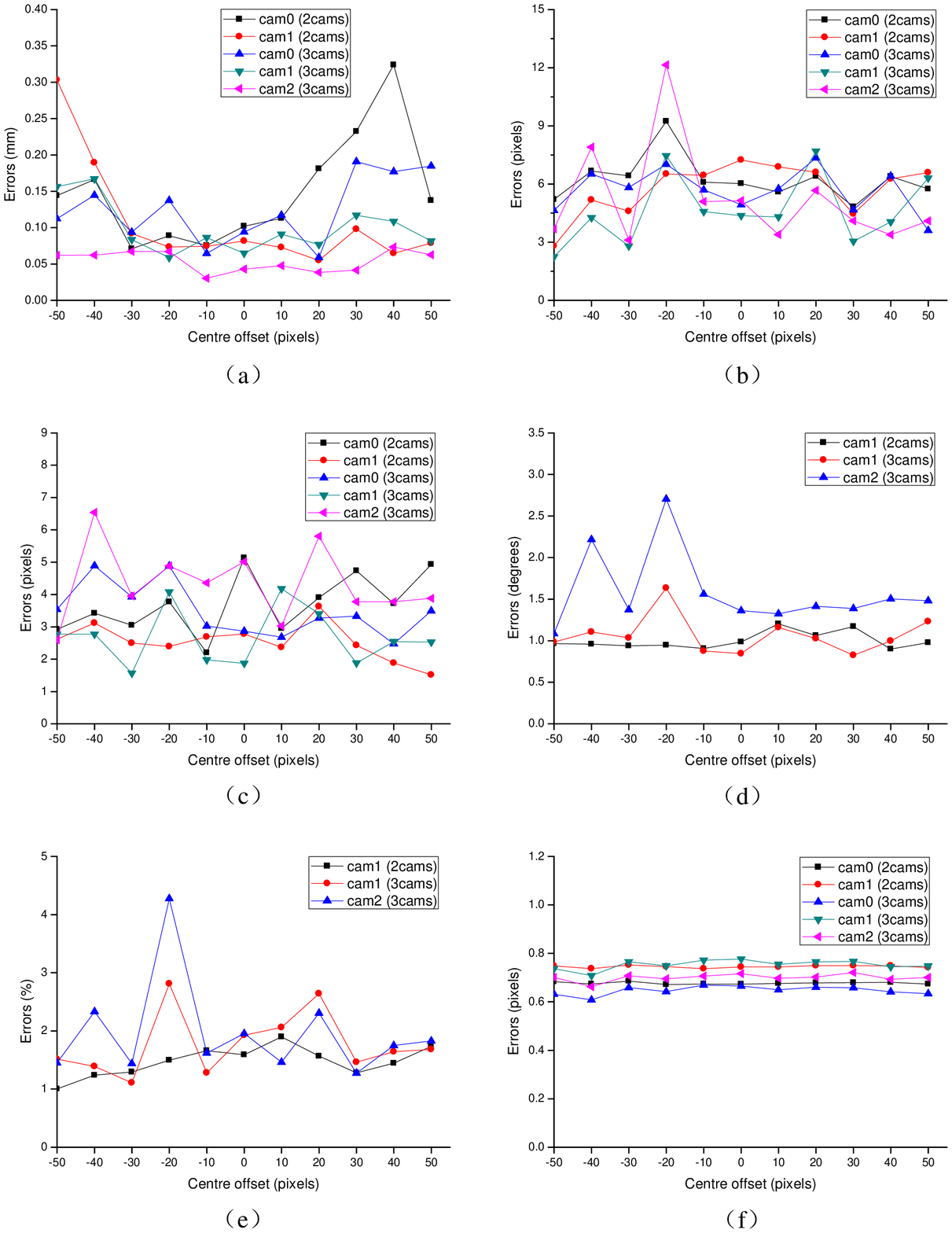}
\end{center}
\par
\vspace{-2em}\caption{Calibration errors of the intrinsic and extrinsic
parameters and reprojection errors for different center offsets of the
principle points. `2cams' means calibration of the $0th,1th$ fish-eye cameras
and `3cams' means calibration of the $0th,1th,2th$ fish-eye cameras: (a) focal
lengths of two and multiple cameras; (b) principle points ($u_{0}$) of two and
multiple cameras; (c) principle points ($v_{0}$) of two and multiple cameras;
(d) calibration error of rotation $E_{r}$ of two and multiple cameras; (e)
calibration error of translation $E_{t}$ of two and multiple cameras; (f) RMS
reprojection error $E_{RMS}$ of two and multiple cameras.}%
\end{figure}

As shown in Fig. 5 and Fig. 6, the calibration errors of the intrinsic and
extrinsic parameters change to a small extent with the focal length offset or
the center offset of the principle point. Moreover, the RMS reprojection
errors of the cameras remain almost constant. In summary, the optimization
always converges to a good solution even when initial solutions largely differ
from the true solution.

\subsection{Real experiments}

In the real experiments, we use Basler scA640-120gm/gc cameras with the image
resolution of 658 pixels $\times$ 492 pixels, equipped with conventional
lenses (Pentax C60402KP) having a FOV of $86.77^{\circ}$ or fish-eye lenses
(Fujinon FE185C057HA-1) having a FOV of $185^{\circ}.$ The nominal focal
lengths of conventional lenses and fish-eye lenses are 4.2 mm and 1.8 mm,
respectively. The 1D calibration wand is a hollow wand with three collinear
LEDs on it (see Fig. 7 (a)) and the distances between the LEDs therein are%
\begin{align*}
L_{1}  &  =\left \Vert A-B\right \Vert =400mm\\
L_{2}  &  =\left \Vert B-C\right \Vert =200mm\\
L  &  =\left \Vert A-C\right \Vert =600mm.
\end{align*}
In order to clearly observe the LEDs, the outside light could be minimized by
setting the exposure time of each camera to a small value. Let the wand
undertake general rigid motion for many times so that image points fill the
image plane as far as possible. Meanwhile, the pixel coordinates of
corresponding image points are obtained by using the geometry of the three
collinear LEDs.

In the following, we investigate the performance of the proposed method and
compare it with the state-of-the-art checkboad-based methods proposed by
Bouguet \cite{Bouguet(2013)} and Kannala \cite{Kannala(2006)}. In this paper,
we use a $7\times10$ checkboard (see Fig. 7 (b)) pattern and the corner points
are detected automatically by using the method in \cite{Geiger(2012)}.
Compared to conventional checkboard-based methods, the proposed method is more
convenient and more efficient especially when there are many cameras to
calibrate. The deficiency of the proposed method is less accurate than
conventional checkboard-based methods, because the feature extraction is less accurate.

First, we perform experiments on two cameras, including two conventional
cameras, two fish-eye cameras and two mixed cameras (camera 0 is a fish-eye
camera and camera 1 is a conventional camera). With some prior knowledge given
by the camera manufacturer and wand constraints, the intrinsic and extrinsic
parameters (including radial distortion parameters) of the two cameras can be
calibrated simultaneously by the proposed algorithm in section III-A. The
calibration results are shown in Tables 1-3, from which we find that the three
methods give similar calibration results.

We also perform experiments on multiple cameras, including two conventional
cameras (camera 0 and camera 2) and a fish-eye camera (camera 1). The two
conventional cameras have a small common FOV, so it is impractical to use
checkboard-based methods to calibrate the intrinsic and extrinsic parameters
of these three cameras simultaneously. However, it is easy to finish this task
by using the proposed algorithm in section III-B. Fig. 8 shows the vision
graph generated from the calibration with camera 0 chosen as the reference
camera. Due to a small overlap between camera 0 and camera 2, the optimal
transformation path for this camera is 0-1-2, rather than 0-2. The calibration
results of intrinsic parameters are shown in Table 4, from which it is found
that the three methods give similar results again.

After calibration, we perform 3D reconstruction with the calibration results
for all the camera setups above. Put the 1D calibration wand randomly at
twenty different places in a measurement volume of 3m$\times$3m$\times$3m.
Thus, for each camera twenty images are taken, samples of which are shown in
Fig. 9. The corresponding pixel coordinates of 3D points $A,C$ are extracted
manually. Tables 1-4 also give the reconstruction results, where%
\begin{equation}
D_{RMS}=\sqrt{\frac{1}{20}\sum_{j=1}^{20}\left(  L-\left \Vert A_{j}^{r}%
-C_{j}^{r}\right \Vert \right)  ^{2}}%
\end{equation}
with $A_{j}^{r},C_{j}^{r}$ being the reconstructed points of $A,C$ for the
$j$th image pair or image triple. We know from these tables that the proposed
method and \cite{Bouguet(2013)} have similar measurement accuracy in the case
of two conventional cameras. However, if there are two fish-eye cameras or two
mixed cameras, then our method gives better results than \cite{Bouguet(2013)}.
This is probably because: i) the 1D calibration wand is freely placed in the
scene volume, and this can increase the calibration accuracy; ii) our 2D
pattern is a printed paper on a board, thus it is not accurate enough. From
Tables 1-4, it is concluded that the measurement error of the proposed method
is about 1\% for all camera setups.%
\begin{gather*}
\text{{\small Table 1. Calibration results of the intrinsic and extrinsic
parameters and reconstruction results of two conventional cameras.}}\\
\text{{\small
\begin{tabular}
[c]{|c|c|c|c|c|c|c|}\hline
Method & \multicolumn{2}{|c}{Proposed} & \multicolumn{2}{|c}{Bouguet} &
\multicolumn{2}{|c|}{Kannala}\\ \hline \hline
Camera & cam 0 & cam 1 & cam 0 & cam 1 & cam 0 & cam 1\\ \hline
$k_{1}$ & \  \  \  \ 4.5932\  \  \  \  & 3.3307 & \  \  \  \  \  \  \ N/A\  \  \  \  \  \  \  &
N/A & 4.3547 & 4.0564\\ \hline
$k_{2}$ & -0.6424 & 0.0200 & N/A & N/A & -0.5023 & 0.2402\\ \hline
$u_{0}$(pixel) & 355.4040 & 376.8750 & 354.8109 & 370.3515 & 343.2948 &
361.6145\\ \hline
$v_{0}$(pixel) & \ 236.2023 & 271.7577 & 230.5293 & 268.0877 & 223.6586 &
293.3133\\ \hline
$R$ & \multicolumn{2}{|c}{$[0.0729,0.6276,0.2053]^{T}$} &
\multicolumn{2}{|c}{$[0.0774,0.6365,0.1997]^{T}$} & \multicolumn{2}{|c|}{N/A}%
\\ \hline
$T$(mm) & \multicolumn{2}{|c}{$[-1373.47,-180.03,504.96]^{T}$} &
\multicolumn{2}{|c}{$[-1371.46,-174.70,529.24]^{T}$} &
\multicolumn{2}{|c|}{N/A}\\ \hline
$E_{RMS}$(pixel) & 0.5817 & 0.5421 & N/A & N/A & N/A & N/A\\ \hline
$D_{RMS}$(mm) & \multicolumn{2}{|c}{6.3160} & \multicolumn{2}{|c}{5.9858} &
\multicolumn{2}{|c|}{N/A}\\ \hline
\end{tabular}
}}%
\end{gather*}

\begin{gather*}
\text{{\small Table 2. Calibration results of the intrinsic and extrinsic
parameters and reconstruction results of two fish-eye cameras.}}\\
\text{{\small
\begin{tabular}
[c]{|c|c|c|c|c|c|c|}\hline
Method & \multicolumn{2}{|c}{Proposed} & \multicolumn{2}{|c}{Bouguet} &
\multicolumn{2}{|c|}{Kannala}\\ \hline \hline
Camera & cam 0 & cam 1 & cam 0 & cam 1 & cam 0 & cam 1\\ \hline
$k_{1}$ & \  \  \  \ 1.8449 \  \  \  & 1.7273 & \  \  \  \ N/A \  \  \  \  & N/A &
1.7558 & 1.7083\\ \hline
$k_{2}$ & -0.0033 & 0.0753 & N/A & N/A & 0.0706 & 0.1061\\ \hline
$u_{0}$(pixel) & \  \  \ 350.2229 \  \  & 352.2591 & \  \  \ 355.6940 \  \  &
357.9337 & 344.8255 & 356.7091\\ \hline
$v_{0}$(pixel) & \ 238.8122 & 256.1896 & \ 236.9247 & 257.9265 & 237.3625 &
248.5101\\ \hline
$R$ & \multicolumn{2}{|c}{$[0.0679,0.7277,0.2314]^{T}$} &
\multicolumn{2}{|c}{$[0.0749,0.7283,0.2272]^{T}$} & \multicolumn{2}{|c|}{N/A}%
\\ \hline
$T$(mm) & \multicolumn{2}{|c}{$[-1277.65,-149.39,475.19]^{T}$} &
\multicolumn{2}{|c}{$[-1276.43,-149.40,452.62]^{T}$} &
\multicolumn{2}{|c|}{N/A}\\ \hline
$E_{RMS}$(pixel) & 0.3948 & 0.3575 & N/A & N/A & N/A & N/A\\ \hline
$D_{RMS}$(mm) & \multicolumn{2}{|c}{5.0890} & \multicolumn{2}{|c}{16.1052} &
\multicolumn{2}{|c|}{N/A}\\ \hline
\end{tabular}
}}%
\end{gather*}

\begin{gather*}
\text{{\small Table 3. Calibration results of the intrinsic and extrinsic
parameters and reconstruction results of two mixed cameras.}}\\
\text{{\small
\begin{tabular}
[c]{|c|c|c|c|c|c|c|}\hline
Method & \multicolumn{2}{|c}{Proposed} & \multicolumn{2}{|c}{Bouguet} &
\multicolumn{2}{|c|}{Kannala}\\ \hline \hline
Camera & cam 0 & cam 1 & cam 0 & cam 1 & cam 0 & cam 1\\ \hline
$k_{1}$ & \  \  \  \ 1.8192 \  \  \  & 4.1128 & \  \  \  \ N/A \  \  \  \  & N/A &
1.7448 & 4.0191\\ \hline
$k_{2}$ & -0.1012 & 0.5818 & N/A & N/A & 0.0235 & -0.5901\\ \hline
$u_{0}$(pixel) & \  \  \ 357.1181 \  \  & 381.2913 & \  \  \ 348.0850 \  \  &
362.4492 & 359.5363 & 360.5308\\ \hline
$v_{0}$(pixel) & \ 237.8219 & 258.8621 & \ 243.3855 & 271.0786 & 240.3054 &
248.9361\\ \hline
$R$ & \multicolumn{2}{|c}{$[0.0649,0.7165,0.2285]^{T}$} &
\multicolumn{2}{|c}{$[0.0743,0.7095,0.2360]^{T}$} & \multicolumn{2}{|c|}{N/A}%
\\ \hline
$T$(mm) & \multicolumn{2}{|c}{$[-1297.01,-149.46,450.24]^{T}$} &
\multicolumn{2}{|c}{$[-1270.74,-150.47,485.44]^{T}$} &
\multicolumn{2}{|c|}{N/A}\\ \hline
$E_{RMS}$(pixel) & 0.3526 & 0.6980 & N/A & N/A & N/A & N/A\\ \hline
$D_{RMS}$(mm) & \multicolumn{2}{|c}{8.3876} & \multicolumn{2}{|c}{12.3939} &
\multicolumn{2}{|c|}{N/A}\\ \hline
\end{tabular}
}}%
\end{gather*}

\begin{gather*}
\text{{\small Table 4. Calibration results of the intrinsic parameters and
reconstruction results of multiple cameras.}}\\
\text{\text{{\small
\begin{tabular}
[c]{|c|c|c|c|c|c|c|c|c|c|}\hline
Method & \multicolumn{3}{|c}{Proposed} & \multicolumn{3}{|c}{Bouguet} &
\multicolumn{3}{|c|}{Kannala}\\ \hline \hline
Camera & cam 0 & cam 1 & cam 2 & cam 0 & cam 1 & cam 2 & cam 0 & cam 1 & cam
2\\ \hline
$k_{1}$ & 3.2958 & 1.7573 & 4.1353 & N/A & N/A & N/A & 4.1519 & 1.6884 &
4.1626\\ \hline
$k_{2}$ & -1.0775 & -0.0872 & -1.6900 & N/A & N/A & N/A & -0.1672 & -0.0177 &
-0.2902\\ \hline
$u_{0}$(pixel) & 338.1441 & 346.9810 & 356.9225 & 324.4709 & 354.5205 &
347.8188 & 324.8254 & 355.0662 & 347.5127\\ \hline
$v_{0}$(pixel) & 231.6928 & 251.2369 & 269.0179 & 228.0510 & 260.3225 &
267.9654 & 228.9251 & 259.4588 & 260.4136\\ \hline
$E_{RMS}$(pixel) & 0.5364 & 0.3148 & 0.5763 & N/A & N/A & N/A & N/A & N/A &
N/A\\ \hline
$D_{RMS}$(mm) & \multicolumn{3}{|c}{3.9091} & \multicolumn{3}{|c}{N/A} &
\multicolumn{3}{|c|}{N/A}\\ \hline
\end{tabular}
}}}%
\end{gather*}
\newline

Note that the methods in \cite{Shah(1997)} and \cite{Nishimoto(2007)} require
parallel stereo vision to perform 3D measurement. However, this is not a
requirement for the proposed method in this paper. In the experiments above,
the differences between the measuring distances and the ground truth may come
from several sources, such as the inaccurate extraction of image points and
the manufacture errors of the 1D calibration wand. Although there are so many
error sources, the calibration accuracy obtained by using the proposed method
is satisfying. Also, these experiments demonstrate the practicability of the
proposed calibration method.

\section{Conclusions}

A calibration method with a one-dimensional object under general motions is
proposed to calibrate multiple fish-eye cameras in this paper. Simulations and
real experiments have demonstrated that the calibration method is accurate,
efficient and user-friendly. The proposed method is generic and also suitable
for two/multiple conventional cameras and mixed cameras. When there are two
conventional cameras, the proposed method and 2D pattern based methods have
similar calibration accuracy. However, the proposed method gives more accurate
calibration results in case of two fish-eye cameras or two mixed cameras (a
conventional camera and a fish-eye camera). The achieved level of accuracy for
two fish-eye cameras is promising, especially considering their use for 3D
measurement purposes. In addition, 2D/3D pattern based methods are
inapplicable when multiple cameras have little or no common FOV, whereas it is
not a problem for the proposed method.

\begin{figure}[h]
\begin{center}
\includegraphics[
scale=0.8]{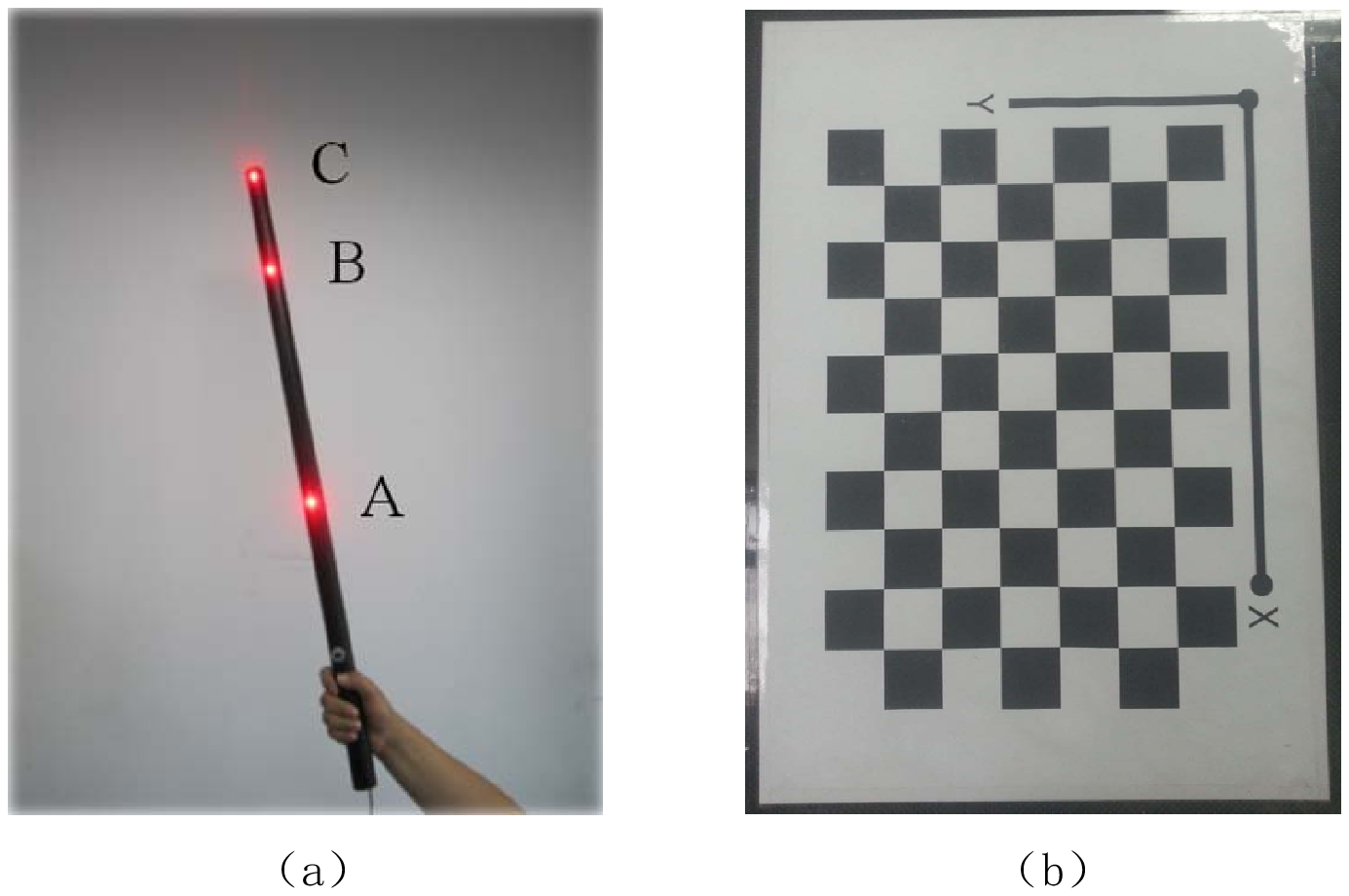}
\end{center}
\par
\vspace{-2em}\caption{Calibration objects: (a) one-dimensional calibration
wand used in the proposed method; (b) checkboard used in the comparison of the
proposed method.}%
\end{figure}\begin{figure}[h]
\begin{center}
\includegraphics[
scale=0.8]{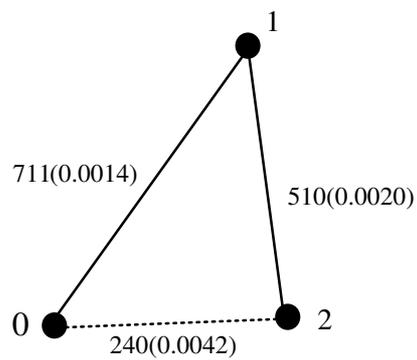}
\end{center}
\par
\vspace{-2em}\caption{Vision graph and the optimal path from reference camera
0 to the other two cameras in solid lines. Numbers indicate the number of
common points between cameras and their corresponding weights.}%
\end{figure}\begin{figure}[h]
\begin{center}
\includegraphics[
scale=0.8]{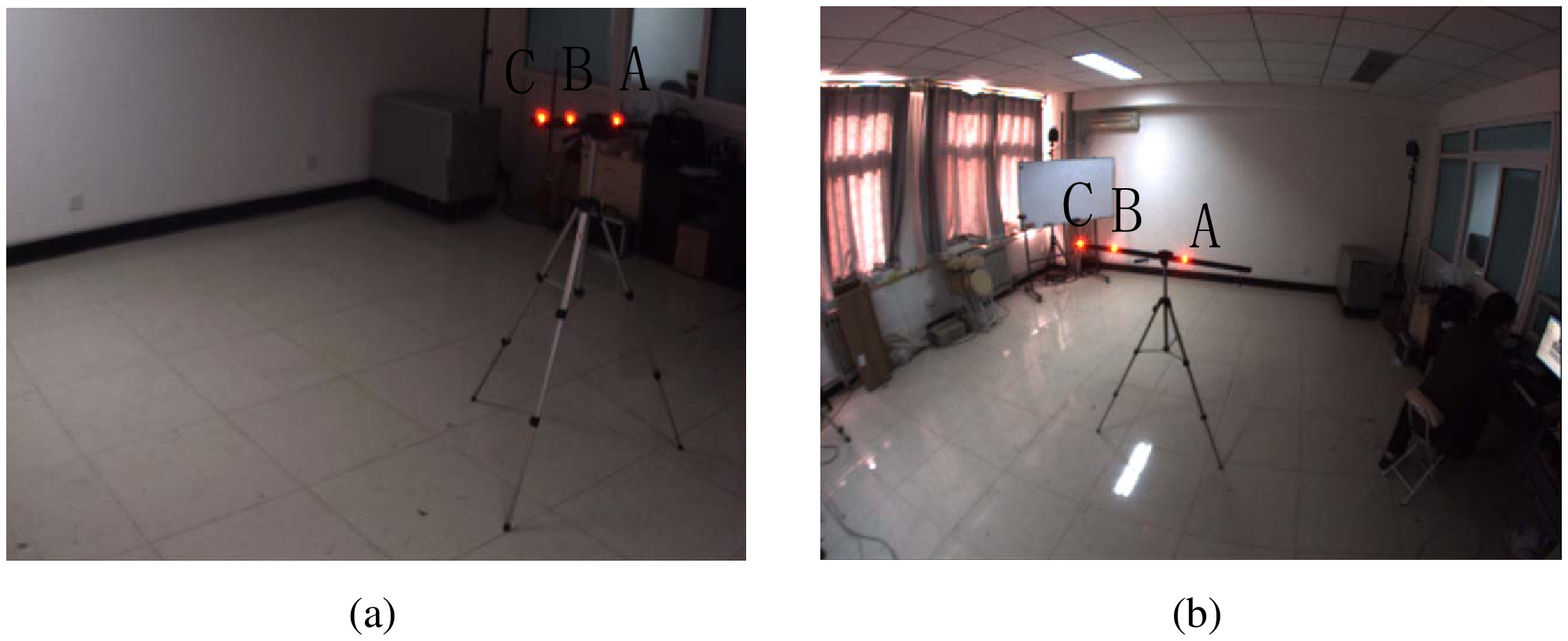}
\end{center}
\par
\vspace{-2em}\caption{Sample images of the 1D object captured by the cameras
for reconstruction: (a) a conventional camera; (b) a fish-eye camera.}%
\end{figure}

\end{document}